\documentclass[twoside,twocolumn,11pt]{article}

\usepackage{fancyhdr}
\usepackage{geometry}
\usepackage{abstract}
\usepackage{graphicx}
\usepackage{titlesec}
\usepackage{ragged2e}
\usepackage{amssymb}
\usepackage{parskip}
\usepackage{amsmath}
\usepackage{newtxtext}
\usepackage{microtype}
\usepackage{booktabs}
\usepackage{array}
\usepackage{balance}
\usepackage[backend=bibtex,style=ieee]{biblatex}
\usepackage[font=footnotesize,labelfont=bf,justification=raggedright,format=plain]{caption}
\usepackage{float}
\usepackage{hyperref}
\usepackage[most]{tcolorbox}
\usepackage{makecell}
\usepackage{tcolorbox}
\tcbuselibrary{skins, breakable}

\newtcolorbox{promptbox}[1][Prompt]{
  enhanced,
  breakable,
  colback=gray!2,
  colframe=gray!30,
  colbacktitle=gray!15,
  coltitle=black,
  boxrule=0.5pt,
  arc=3pt,
  outer arc=3pt,
  lefttitle=8pt,
  toptitle=5pt,
  bottomtitle=5pt,
  left=8pt,
  right=8pt,
  top=7pt,
  bottom=7pt,
  fonttitle=\bfseries\small,
  title=#1,
  attach boxed title to top left={yshift=0pt},
  boxed title style={
    colback=gray!22,
    colframe=gray!30,
    boxrule=0.5pt,
    arc=3pt,
    outer arc=3pt,
  },
  separator sign={},
}

\usepackage{colortbl}
\definecolor{fablerow}{gray}{0.91}

\usepackage{colortbl}
\usepackage{arydshln}
\definecolor{verdictgreen}{RGB}{0, 120, 80}
\definecolor{verdictorange}{RGB}{180, 90, 0}
\definecolor{fablerow}{gray}{0.91}

\usepackage{tikz}
\usetikzlibrary{shapes.geometric, arrows.meta, positioning}

\defbibheading{bibliography}[\refname]{%
  \footnotesize\section{\MakeUppercase{REFERENCES}}}

\newcolumntype{L}[1]{>{\raggedright\arraybackslash}p{#1}}
\newcolumntype{C}[1]{>{\centering\arraybackslash}p{#1}}
\newcolumntype{R}[1]{>{\raggedleft\arraybackslash}p{#1}}

\fancypagestyle{firstpage}{
  \fancyhf{}
  \fancyhead[LO]{\small}
  \fancyhead[RO]{\small}
  \fancyfoot[L]{\footnotesize
    \textsuperscript{*}Corresponding Author: Dominic Okonkwo \textless dominic.okonkwo@uga.edu\textgreater\\
    \textit{All models in this study were evaluated single-shot; few-shot conditions were not run due to the token cost of the models evaluated, and remain an open direction for future work.}
  }
  \fancyfoot[R]{}

}

\title{\fontsize{16}{20}\selectfont\textbf{Capabilities of Claude Fable 5 on Biomedical Challenge Problems}}

\author{
    \fontsize{12}{14}\selectfont
    Dominic Okonkwo\textsuperscript{1*},
    Magnus Hodgson\textsuperscript{1},
    Temitope I. David\textsuperscript{2},
    Susan Adanna Ihejirika\textsuperscript{3}\\[1ex]
    \fontsize{10}{12}\selectfont
    \textsuperscript{1}School of Computing, University of Georgia \\
    \fontsize{10}{12}\selectfont
    \textsuperscript{2}Department of Chemistry, University of Illinois \\
    \fontsize{10}{12}\selectfont
    \textsuperscript{3}Institute of Bioinformatics, University of Georgia
}

\date{}

\titleformat{\section}
{\normalsize\bfseries}
{\thesection.}
{1em}
{}

\titleformat{\subsection}
{\normalsize\bfseries\flushleft}
{\thesubsection.}
{1em}
{}

\titleformat{\subsubsection}
{\normalsize\bfseries\itshape\flushleft}
{\thesubsubsection.}
{1em}
{}

\titlespacing*{\section}{0pt}{8pt}{4pt}
\titlespacing*{\subsection}{0pt}{8pt}{4pt}
\titlespacing*{\subsubsection}{0pt}{3pt}{1pt}
\begin{document}

\twocolumn[
\begin{@twocolumnfalse}

\vspace{-0.7cm}
\noindent
\raggedright
{\fontsize{11}{13}\selectfont}
\hfill
{\fontsize{11}{13}\selectfont\footnotesize }

\vspace{-0.5cm}

\maketitle
\thispagestyle{firstpage}

\begin{center}
{\normalsize\bfseries Abstract}
\end{center}

\vspace{0.3em}

\begin{center}
\begin{minipage}{0.94\textwidth}
\fontsize{10}{12}\selectfont
\justifying

Frontier language models are increasingly evaluated on biomedical benchmarks, but two problems undermine most published evaluations: legacy benchmarks are near-saturated, and open-ended responses are graded by other language models. We evaluate Claude Fable 5, Anthropic's most capable publicly available model, across eight biomedical benchmarks, four text and four multimodal, using deterministic scoring against fixed answer keys throughout. We include two Claude predecessors and GPT-5 as baselines. Refusal is tracked as a distinct outcome in every result table. That decision produces the paper's central finding. Fable 5 refuses between 8.0\% and 99.4\% of questions depending on the benchmark, a pattern absent in both predecessors and in GPT-5. Once refused items are excluded from the denominator, Fable 5's accuracy exceeds or meets every other model on every benchmark in this study. We identify two distinguishable refusal patterns: one concentrating in basic-science and mechanism content across MedQA and MedXpertQA MM, confirmed independently on two benchmarks using each benchmark's own category labels; and a separate disease-domain pattern on RareBench, where inborn metabolic disease presentations are refused near-universally while adult-onset autoimmune presentations are not. The primary constraint on Fable 5's biomedical usefulness is willingness to engage, not capability once it does.
\end{minipage}
\end{center}

\vspace{0.8cm}

\setcounter{figure}{0}
{\centering\noindent\includegraphics[width=0.75\textwidth]{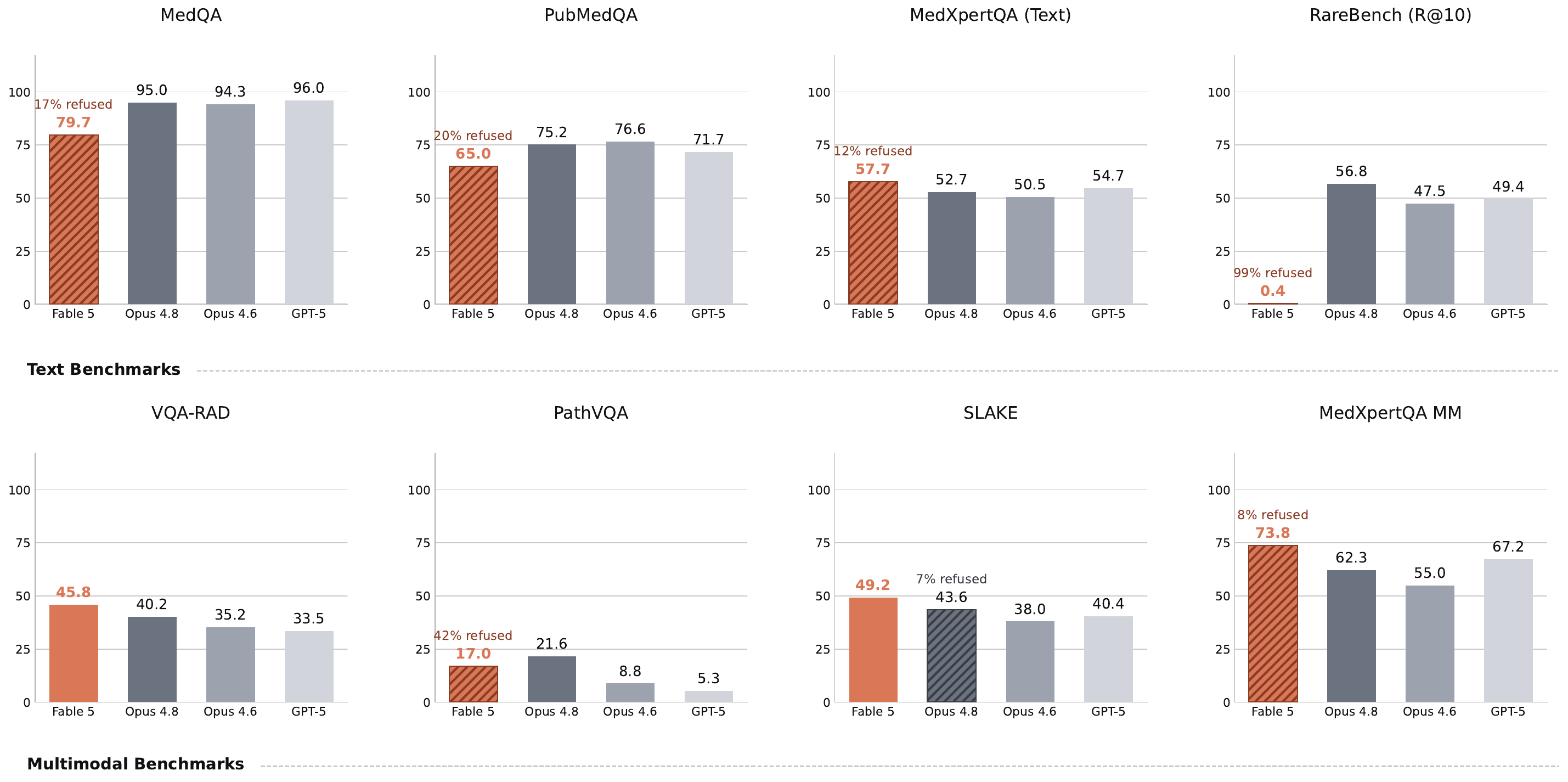}\par}

\vspace{2pt}

{\footnotesize Overview of the eight biomedical benchmarks evaluated in this study. Raw accuracy across eight biomedical benchmarks; refusal rate annotated in red where non-negligible ($\geq$5\%). \par}

\vspace{0.8cm}

\end{@twocolumnfalse}
]


\section{Introduction}
\fontsize{11}{13}\selectfont

Biomedical evaluation of language models has not kept pace with the models themselves. Every new frontier model claims stronger scientific reasoning, and every new model report leads with a wall of benchmark scores. But two specific practices have quietly undermined what those scores mean. First, models have gotten good enough that many established exams no longer separate one system from the next, with frontier models now routinely exceeding 90\% accuracy on USMLE-style examinations \cite{Vishwanath2026GeneralPurposeLLMs}, once every model scores the same, the benchmark stops telling you which one is actually better. Second, grading is increasingly outsourced to other language models, which means the evaluation is only as trustworthy as the judge doing the grading \cite{Gu2024LLMJudge}. Neither problem means existing benchmarks are without value; it means using them well requires choosing carefully which ones still have room to discriminate between models, and scoring them in a way that does not introduce a second, unaccountable model into the judgment.

This is especially true in medicine and biology, where the gap between ``sounds right'' and ``is right'' matters more than almost anywhere else. A wrong answer on a trivia benchmark is a curiosity. A wrong answer about a rare disease, a drug interaction, or a lab image is not. Evaluating a model in this domain has to do more than confirm it can pass an exam that was hard five years ago. It has to test reasoning that is still difficult today, refuse to let another language model quietly decide what counts as correct, and take seriously that real biomedical questions increasingly come with an image attached, a scan, a slide, a chart, not just text \cite{Zuo2025MedXpertQA}.

Claude Fable 5 is the subject of this paper. It is Anthropic's most capable publicly available model, trained with large-scale reinforcement learning and positioned as a step forward in scientific and medical reasoning. Released on June 9, 2026, with public API access restored July 1, 2026, its biomedical capabilities have not been independently evaluated in the published literature. This paper provides that evaluation, following the zero-shot, direct-answer evaluation template established for frontier models on medical benchmarks \cite{Nori2023GPT4Medical}, and adds two constraints most evaluations skip: \textbf{every benchmark is scored against a fixed answer key, never by another model's judgment}, and \textbf{refusing to answer is tracked and reported as its own outcome}, not quietly folded into incorrect. Our primary evidence benchmarks were chosen specifically because current frontier models have not already maxed them out; two further benchmarks are included despite being near-ceiling, precisely because their saturation lets us measure how far the field has moved since they were first used.

Most evaluations treat a refusal the same as a wrong answer. \textbf{This one does not.} Refusal behavior in large language models, where a model declines to respond rather than producing an answer, is increasingly understood as a distinct model output class, shaped by post-training alignment rather than by capability alone \cite{Joad2026RefusalDirection}. A model that declines to answer and a model that answers confidently and incorrectly are not the same failure, and collapsing that distinction discards exactly the information that matters most when deciding whether to deploy a model in a real clinical or research setting. That decision turns out to produce this paper's central finding, one that appears not just on the benchmark most likely to trigger it, but across nearly every dataset in this study, including two benchmarks we included only for historical comparability and expected to be unremarkable.

This work makes four contributions. First, a text evaluation spanning MedQA \cite{Jin2021MedQA}, PubMedQA \cite{Jin2019PubMedQA}, MedXpertQA \cite{Zuo2025MedXpertQA}, and RareBench \cite{Chen2024RareBench}, the first two retained for comparability to prior work despite being near-saturated, the latter two chosen because current models still visibly struggle with both. Second, a multimodal evaluation using the harder, open-ended portions of VQA-RAD \cite{Lau2018VQARAD}, SLAKE \cite{Liu2021SLAKE}, PathVQA \cite{He2020PathVQA}, and the MedXpertQA multimodal subset \cite{Zuo2025MedXpertQA}, skipping the closed-ended subsets that no longer discriminate among frontier systems. Third, a comparison against GPT-5 \cite{OpenAI2026GPT5SystemCard} under matched prompting and scoring conditions. Fourth, a generational trace within a single model family, Claude Opus 4.6, Opus 4.8, and Fable 5, evaluated under an identical protocol, documenting whether biomedical capability is actually improving release to release, or only appearing to.

\textbf{Refusals are reported as a distinct outcome in every result in this paper}, not folded silently into an accuracy score. What those refusals mean, why they occur, and what they reveal about where Fable 5 holds up under pressure and where it does not, that is what this paper is for. Anyone deciding whether to trust it with a real biomedical question deserves that answer plainly, not buried inside a leaderboard number.

\section{Methodology}
\label{sec:methodology}
\fontsize{11}{13}\selectfont

\subsection{Models}
\label{sec:models}

We evaluate four models: Claude Fable 5 (the primary subject), Claude Opus 4.6 and Opus 4.8 as within-family predecessors, and GPT-5 from OpenAI as an external baseline. We access all Claude models via the Anthropic Messages API and GPT-5 via the OpenAI Chat Completions API. All four models are vision-capable, which the multimodal phase requires.

\begin{table}[h]
\centering
\footnotesize
\renewcommand{\arraystretch}{1.22}
\setlength{\tabcolsep}{5pt}
\begin{tabular}{p{2.0cm} c p{3.5cm}}
\toprule
\textbf{Benchmark} & \textbf{\textit{n}} & \textbf{Task / Domain} \\
\midrule
\multicolumn{3}{l}{\textit{\textbf{Text Benchmarks}}} \\[2pt]
MedQA        & 1,273 & USMLE-style clinical MCQ \\
PubMedQA     & 1,000 & Biomedical literature QA \\
MedXpertQA   & 2,450 & Specialist board exam QA \\
RareBench    & 1,122 & Rare disease diagnosis \\[4pt]
\multicolumn{3}{l}{\textit{\textbf{Multimodal Benchmarks}}} \\[2pt]
MedXpertQA MM & 400  & Specialist clinical images \\
VQA-RAD      &   179 & Radiology VQA \\
SLAKE        &   250 & Radiology VQA \\
PathVQA      &   171 & Histopathology VQA \\
\bottomrule
\end{tabular}
\caption{Biomedical benchmarks used in this study.}
\label{tab:benchmark_summary}
\vspace{-8pt}
\end{table}

\subsection{Datasets}

We evaluate across two phases, text-only and multimodal, covering eight benchmarks in total. Table~\ref{tab:benchmark_summary} summarizes each benchmark, its size, domain, and role in this study.
For the text phase, MedQA and PubMedQA serve as legacy anchors; MedXpertQA Text and RareBench serve as primary evidence benchmarks.
For the multimodal phase, we sample 1,000 items across four image-based benchmarks selected to span radiology, multi-modality imaging, and histopathology, using a fixed random seed, drawing only from open-ended subsets, which retain more headroom than closed-ended items. MedXpertQA MM uses a specialist-level, five to ten-option format across 17 specialties, distinct in design from the shorter-answer VQA-RAD, SLAKE, and PathVQA tasks. We use the complete available open-ended test set for VQA-RAD, filter SLAKE to English-language open-ended questions, and subsample PathVQA from a larger histopathology pool for cost control.

\subsection{Prompting}
\label{sec:prompting}

We conduct all evaluations under zero-shot, direct-answer prompting. We do not use few-shot examples, chain-of-thought instruction, retrieval augmentation, or extended thinking anywhere in the pipeline. Each model receives a single user-turn message with no system prompt.

For MCQ benchmarks, we instruct models to place their final answer on the last line in the format \texttt{"ANSWER: <letter>"}. For open-ended VQA benchmarks, we instruct models to answer briefly, a word or short phrase, in the format \texttt{"ANSWER: <answer>"}. For PubMedQA, we prepend the abstract context and instruct the model to answer "yes, no, or maybe" using only that context. For RareBench, we convert Human Phenotype Ontology (HPO) codes \cite{Kohler2019HPOExpansion} to human-readable symptom names and ask for ten ranked candidate diagnoses under the following prompt shown here.

Following the discovery of near-total refusal under this prompt (Section~\ref{sec:rarebench}), we ran a secondary neutral-framing condition on all originally-refused items to test whether the clinical role framing was a contributing factor.

\begin{figure}[t]
\begin{center}
\begin{minipage}{0.8\columnwidth}
\begin{promptbox}[RareBench - Primary Prompt]
\small
You are a clinical decision support system assisting a specialist physician. The following phenotype features have been recorded for a case under review in a rare disease diagnostic workup:

\medskip
\texttt{\{symptom\_str\}}
\medskip

Based on this phenotype profile, provide the top 10 rare disease diagnoses most consistent with these features, ranked from most to least likely. End your response with a numbered list in exactly this format:

\medskip
\texttt{TOP10:}\\
\texttt{1.\ <disease name>}\\
\texttt{2.\ <disease name>}\\
\texttt{3.\ \ldots}\\
\texttt{4.\ <disease name>}
\end{promptbox}
\end{minipage}
\end{center}
\end{figure}

\begin{figure}[t]
\begin{center}
\begin{minipage}{0.88\columnwidth}
\begin{promptbox}[RareBench - Neutral Prompt] 
\small
A patient has the following clinical features:

\medskip
\texttt{\{symptom\_str\}}
\medskip

List the ten most likely diagnoses, ranked from most to least likely. Respond in exactly this format:

\medskip
\texttt{TOP10:}\\
\texttt{1.\ <diagnosis>}\\
\texttt{2.\ <diagnosis>}\\
\texttt{3.\ \ldots}\\
\texttt{4.\ <diagnosis>}
\end{promptbox}
\end{minipage}
\end{center}
\vspace{-8pt}
\end{figure}

This condition is a robustness check on the refusal finding only and does not replace the primary-prompt results reported for RareBench anywhere in this paper.

\subsection{Image Handling}

For the multimodal phase, we load each image from disk, resize if either dimension exceeds 1,568 pixels, and encode as base64 JPEG. We pass images to Claude models as base64 content blocks preceding the text prompt, and to GPT-5 as inline base64 data URIs. For MedXpertQA MM cases with multiple images, we submit all images for that case together in a single API call, in the order given by the dataset metadata.

\begin{table*}[t]
\centering
\label{tab:text_results}
\renewcommand{\arraystretch}{1.35}
\setlength{\tabcolsep}{7pt}
\resizebox{\textwidth}{!}{
\begin{tabular}{l cc cc cc}
\toprule
& \multicolumn{2}{c}{\textbf{MedQA}}
& \multicolumn{2}{c}{\textbf{PubMedQA}}
& \multicolumn{2}{c}{\textbf{MedXpertQA Text}} \\
\cmidrule(lr){2-3} \cmidrule(lr){4-5} \cmidrule(lr){6-7}
\textbf{Model}
  & \textit{Raw}
  & \textit{Scored}
  & \textit{Raw}
  & \textit{Scored}
  & \textit{Raw}
  & \textit{Scored} \\
\midrule
\rowcolor{fablerow}
\textbf{Fable 5}
  & 79.7\%
  & \textbf{96.6\%} {\scriptsize[95.3, 97.5]}
  & 65.0\%
  & \textbf{81.3\%} {\scriptsize[78.4, 83.8]}
  & 57.7\%
  & \textbf{66.1\%} {\scriptsize[64.0, 68.0]} \\
Opus 4.8
  & 95.0\% {\scriptsize[93.6, 96.0]}
  & {\small\textit{-- (0.4\% ref.)}}
  & 75.2\% {\scriptsize[72.4, 77.8]}
  & {\small\textit{-- (0.0\% ref.)}}
  & 52.7\% {\scriptsize[50.8, 54.7]}
  & {\small\textit{-- (0.3\% ref.)}} \\
Opus 4.6
  & 94.3\% {\scriptsize[92.9, 95.5]}
  & {\small\textit{-- (0.2\% ref.)}}
  & 76.6\% {\scriptsize[73.9, 79.1]}
  & {\small\textit{-- (0.0\% ref.)}}
  & 50.5\% {\scriptsize[48.6, 52.5]}
  & {\small\textit{-- (0.1\% ref.)}} \\
GPT-5
  & 96.0\% {\scriptsize[94.8, 96.9]}
  & {\small\textit{-- (0.0\% ref.)}}
  & 71.7\% {\scriptsize[68.8, 74.4]}
  & {\small\textit{-- (0.0\% ref.)}}
  & 54.6\% {\scriptsize[52.7, 56.6]}
  & {\small\textit{-- (0.0\% ref.)}} \\
\bottomrule
\end{tabular}
}
\caption{Text benchmark performance. Scored accuracy excludes refused items from the denominator. 95\% Wilson CIs in brackets.}
\label{tab:text_results}
\end{table*}


\begin{table*}[t]
\centering
\label{tab:rarebench_results}
\renewcommand{\arraystretch}{1.35}
\setlength{\tabcolsep}{8pt}
\resizebox{0.88\textwidth}{!}{
\begin{tabular}{l c c c c c c}
\toprule
\textbf{Model}
  & \textbf{\textit{n}}
  & \textbf{Refused}
  & \textbf{Scored \textit{n}}
  & \textbf{R@1}
  & \textbf{R@5}
  & \textbf{R@10} \\
\midrule
\rowcolor{fablerow}
\textbf{Fable 5}
  & 1,122
  & \textbf{1,115} {\scriptsize(99.4\%)}
  & 6
  & 0.36\% {\scriptsize[0.1, 0.9]}
  & 0.45\% {\scriptsize[0.2, 1.0]}
  & 0.45\% {\scriptsize[0.2, 1.0]} \\
Opus 4.8
  & 1,122
  & 0 {\scriptsize(0.0\%)}
  & 1,122
  & 27.4\% {\scriptsize[24.8, 30.0]}
  & 40.5\% {\scriptsize[37.6, 43.4]}
  & 56.8\% {\scriptsize[53.9, 59.6]} \\
Opus 4.6
  & 1,122
  & 0 {\scriptsize(0.0\%)}
  & 1,122
  & 20.9\% {\scriptsize[18.6, 23.3]}
  & 34.1\% {\scriptsize[31.3, 36.9]}
  & 47.5\% {\scriptsize[44.6, 50.4]} \\
GPT-5
  & 1,122
  & 0 {\scriptsize(0.0\%)}
  & 1,079
  & 27.7\% {\scriptsize[25.2, 30.4]}
  & 41.0\% {\scriptsize[38.2, 43.9]}
  & 49.4\% {\scriptsize[46.5, 52.3]} \\
\bottomrule
\end{tabular}
}
\caption{RareBench performance under the primary prompt. Fable~5 refused 99.4\% of items; accuracy figures reflect the 6 scored responses only.}
\label{tab:rare_text_results}
\end{table*}

\subsection{Scoring and Refusal Handling}

We score all benchmarks deterministically against a fixed answer key; no benchmark in this study uses LLM-based judgment. We report 95\% Wilson confidence intervals on every percentage. Every API call logs the raw response text, API-reported model identifier, stop reason, token counts, and any error string before scoring occurs; we treat the API-reported model identifier as a best-effort signal rather than a confirmed record of internal routing.

For RareBench, we match predicted diagnosis names against a canonical name dictionary via exact match, then substring containment, then fuzzy matching (cutoff 0.6), and report Recall@1, Recall@5, and Recall@10. For the open-ended VQA benchmarks, we score a normalized prediction as correct on exact match or substring containment against the ground-truth short answer.

We flag a response as a refusal if the API stop reason indicates refusal or the response text matches a fixed set of refusal keyword patterns, applied identically across all models and all eight benchmarks. We record refusals as a distinct outcome and exclude them from the correctness denominator in every result table. We report both raw accuracy, refusals counted against the model, and scored-subset accuracy, refusals excluded from the denominator, wherever a model's refusal rate is non-negligible.

\section{Results}
\label{sec:results}
\fontsize{11}{13}\selectfont

\subsection{Refusal Suppresses Fable 5's Raw Scores}
\label{sec:refusal_raw}

\begin{figure}[h]
\centering
\includegraphics[width=0.47\textwidth]{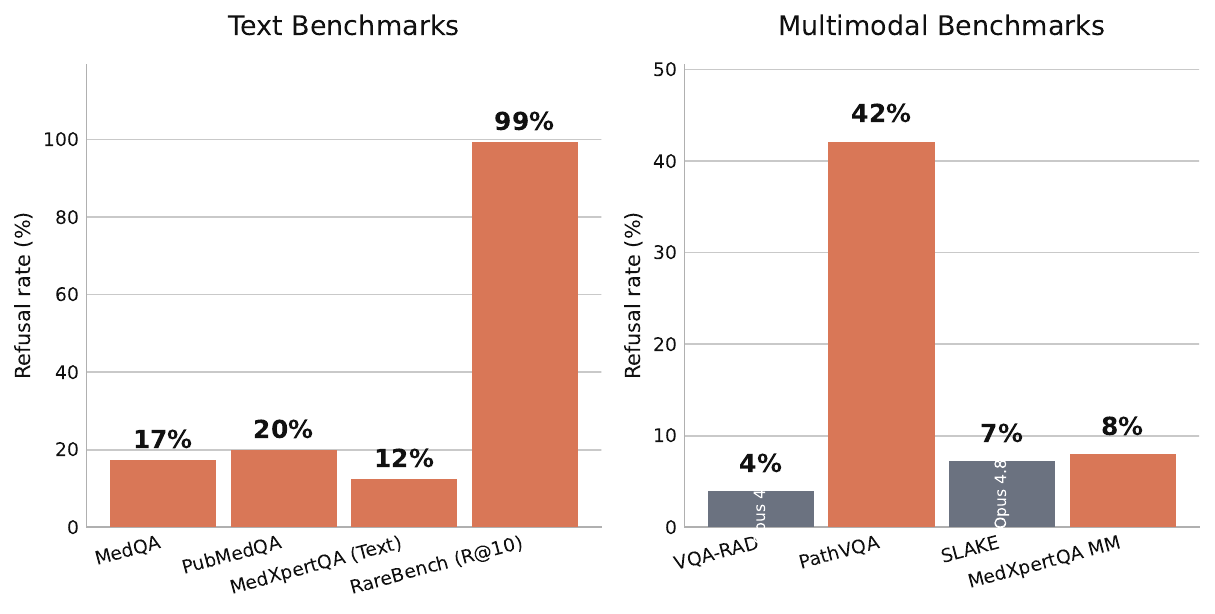}
\caption{Refusal rates across all evaluated biomedical benchmarks.}
\label{fig:refusal_rates_across}
\vspace{-8pt}
\end{figure}

Across all eight benchmarks, Fable 5's raw accuracy understates its capability. The other three models, Opus 4.6, Opus 4.8, and GPT-5, refuse between 0.0\% and 0.4\% of questions on the text benchmarks and show comparable rates on most multimodal benchmarks. Fable 5 does not follow this pattern. It refuses 17.4\% of MedQA (222 of 1,273), 20.0\% of PubMedQA (200 of 1,000), 12.4\% of MedXpertQA Text (305 of 2,450), 41.5\% of PathVQA (71 of 171), and 8.0\% of MedXpertQA MM (32 of 400). Figure ~\ref{fig:refusal_rates_across} shows refusal rates across all eight benchmarks. On RareBench, refusal is the primary outcome; we detail it separately in Section~\ref{sec:rarebench}.

\begin{figure}[h]
\centering
\includegraphics[width=0.47\textwidth]{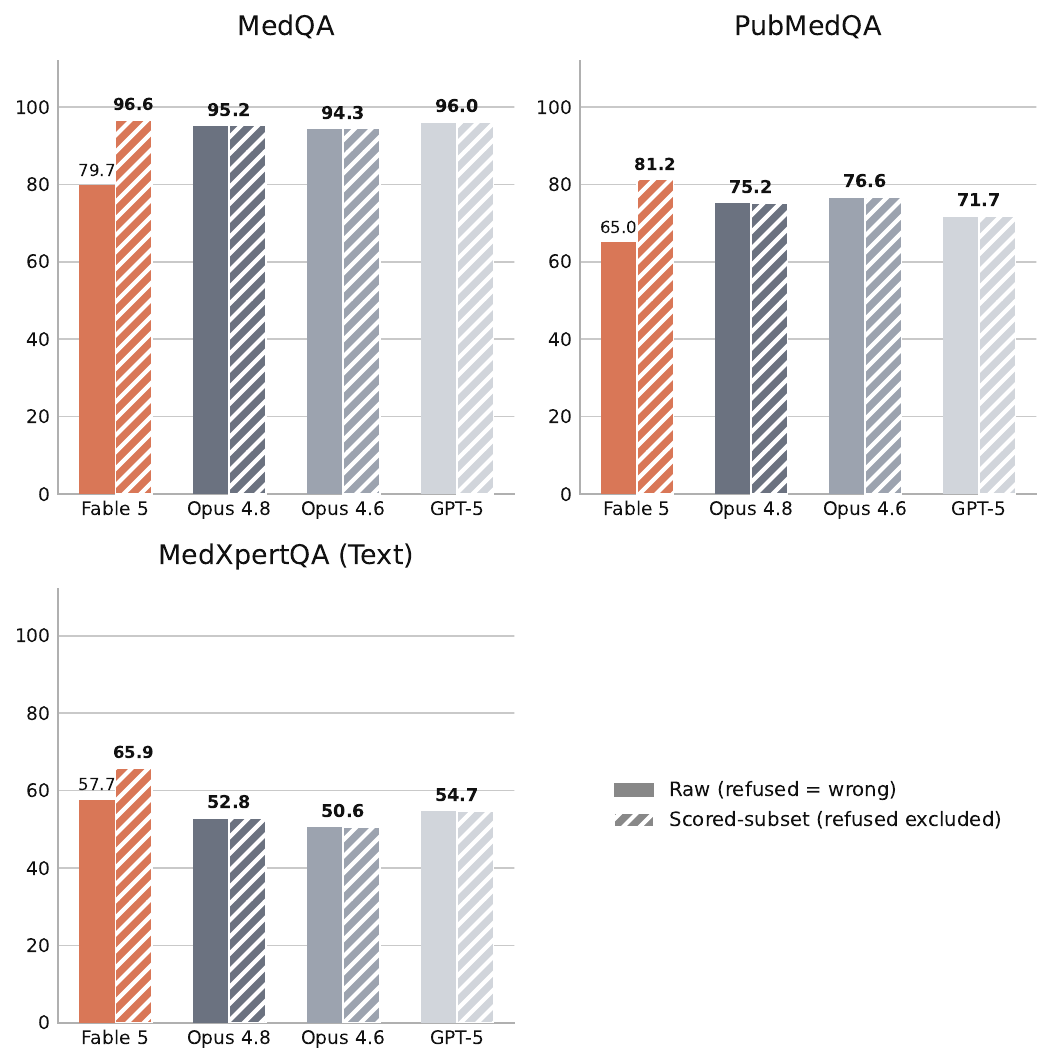}
\caption{Comparison of raw accuracy (refusals counted as incorrect) and scored-subset accuracy (refusals excluded from the denominator) across biomedical benchmarks.}
\label{fig:comparison_rates}
\vspace{-6pt}
\end{figure}

Once refusal is separated from correctness, Fable 5's ranking on every benchmark where its refusal rate is non-negligible moves from last among the four models to leading.
Table~\ref{tab:text_results} reports both raw accuracy and scored-subset accuracy, computed only over items each model actually answered for the text benchmarks. On MedQA, Fable 5's scored-subset accuracy (96.6\% {\scriptsize[95.3, 97.5]}) matches or exceeds every other model's raw accuracy (94.3--96.0\%). On PubMedQA, its scored-subset accuracy (81.3\% {\scriptsize[78.4, 83.8]}) exceeds every other model's raw accuracy (71.7--76.6\%) by a clear margin. On MedXpertQA Text, its scored-subset accuracy (66.1\% {\scriptsize[64.0, 68.0]}) extends what was already a lead in raw terms. There is no benchmark in this study on which Fable 5's demonstrated accuracy, restricted to items it answered, trails the other three models. Figure ~\ref{fig:comparison_rates} plots raw and scored-subset accuracy side by side for the text benchmarks.

The items Fable 5 refused were not uniformly hard. On MedQA, 215 of its 258 total misses (83.3\%) were questions that all three other models answered correctly, meaning the bulk of what suppressed Fable 5's raw MedQA score were questions with no difficulty basis for being missed. This selectivity weakens on PubMedQA (34.6\% uniquely missed) and further on MedXpertQA Text (13.0\%), where Fable 5's remaining misses are shared with the other models and look more like genuinely hard items.

\subsubsection{Refusal does not route to a fallback model}

Anthropic's release documentation for Fable 5 describes triggered queries as falling back to Claude Opus 4.8, which then handles the request normally. If this mechanism were operating in our pipeline, the API's model-identification field should report Opus 4.8 on refused items. It does not. Across all 1,945 refused items logged across every benchmark in both phases, \texttt{model\_reported} returns \texttt{claude-fable-5} in every case, 100.0\%, with zero exceptions. Of the 305 refused MedXpertQA Text responses, 266 (87.2\%) return empty response text alongside the refusal stop reason, consistent with a pre-generation block; the remaining 39 (12.8\%) contain substantial clinical reasoning that is nonetheless blocked rather than returned. Neither pattern is consistent with routing to a different model; both are consistent with a block applied to Fable 5's own output.

\begin{figure}[h]
\centering
\includegraphics[width=0.4\textwidth]{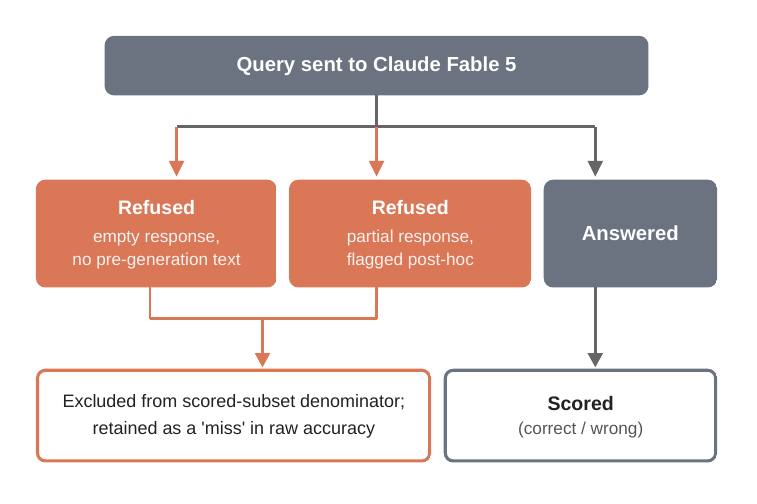}
\caption{Refusal outcome flow, from query submission to final scoring status.}
\label{fig:refusal_outcome}
\end{figure}

\subsection{RareBench: refusal as the primary outcome}
\label{sec:rarebench}

On RareBench, Fable 5 refused 1,115 of 1,122 items (99.4\%) under the study's standard prompt, leaving 6 scoreable responses after excluding one additional unparsed case. Every refusal returned a hard, pre-generation block at the API level, \texttt{stop\_reason: refusal} with empty response text, not a hedged decline. Figure ~\ref{fig:refusal_outcome} shows the full refusal outcome flow from query submission to final scoring status. The other three models refused none. Opus 4.8, Opus 4.6, and GPT-5 each engaged with the full task, with Recall@1 ranging from 20.9\% to 27.7\%, consistent with prior published ranges for non-retrieval-augmented models on this benchmark. Fable 5's Recall@1 of 0.36\% {\scriptsize[0.1, 0.9]}, computed on 6 responses, is not a measure of its diagnostic capability, it is a measure of what survives a near-total refusal wall. Table~\ref{tab:rare_text_results} reports the full RareBench results under the primary prompt

\begin{table*}[t]
\centering
\renewcommand{\arraystretch}{1.45}
\setlength{\tabcolsep}{8pt}
\resizebox{\textwidth}{!}{
\begin{tabular}{p{3.5cm} p{4.8cm} p{5.8cm} p{3.1cm}}
\toprule
\textbf{Hypothesis} & \textbf{Test} & \textbf{Result} & \textbf{Verdict} \\
\midrule

Mortality-related phenotype content &
Refusal rate, cases with vs.\ without mortality-coded HPO term &
100.0\% {\scriptsize(\textit{n}=621)} vs.\ 98.6\% {\scriptsize(\textit{n}=501)} & 
\makecell[tl]{\color{verdictgreen}\textbf{Ruled out}} \\

\hdashline[0.8pt/4pt]

Prompt framing &
Neutral-framing re-run {\scriptsize(\textit{n}=25)} &
92.0\% still refused, 8.0\% flipped &
\makecell[tl]{\color{verdictorange}\textbf{Mostly ruled out}} \\

\hdashline[0.8pt/4pt]

Phenotype-list length &
Flipped vs.\ still-refused items &
Mean 15.0 {\scriptsize(\textit{n}=2)} vs.\ 12.6 {\scriptsize(\textit{n}=23)};\ \textit{p}=0.86 &
\makecell[tl]{\color{verdictgreen}\textbf{Ruled out}} \\

\hdashline[0.8pt/4pt]

\rowcolor{fablerow}
Source subset / disease domain &
Refusal rate by subset &
RAMEDIS 99.7\%, LIRICAL 85.4\%, MME 85.0\%, HMS 48.8\% &
{\textbf{Strongest signal}} \\

\hdashline[0.8pt/4pt]

Mechanism vs.\ presentation content &
Reclassification by phenotype type &
100.0\% vs.\ 99.4\% — no discriminating power &
\makecell[tl]{\color{verdictgreen}\textbf{Ruled out}} \\

\bottomrule
\end{tabular}
}
\caption{Hypotheses tested for the RareBench refusal pattern. The source subset / disease domain hypothesis (highlighted) produced the strongest explanatory signal; all others were ruled out or showed only marginal effects.}
\label{tab:rarebench_hypotheses}
\end{table*}

\subsection{What drives the RareBench refusals}
\label{sec:rarebench_drivers}

We tested five explanations for the refusal pattern. The strongest signal is disease domain. Refusal rate under the original prompt varies sharply by source subset: RAMEDIS 99.7\% (n=624), MME 85.0\% (n=40), LIRICAL 85.4\% (n=369), and HMS 48.8\% (n=82). Manual review of the HMS cases answered under the original prompt shows they are drawn almost entirely from adult-onset autoimmune and rheumatologic presentations, granulomatosis with polyangiitis, Behçet disease, systemic lupus erythematosus. A matched review of refused RAMEDIS cases shows near-uniform representation of severe pediatric inborn metabolic disorders, phenylketonuria, glutaric acidemia, methylmalonic acidemia, the majority annotated with infancy- or childhood-mortality phenotype codes. Four other hypotheses, mortality-related phenotype content, prompt framing, phenotype-list length, and mechanism versus clinical-presentation content, each failed to explain the pattern and are summarized in Table~\ref{tab:rarebench_hypotheses}.

The subset-level pattern is not fully deterministic,  that is, it predicts refusal reliably on average but not for every individual case. LIRICAL contains three Marfan syndrome cases, two answered, one refused, identical diagnosis, same subset, opposite outcomes. The most accurate summary is that refusal correlates strongly with disease domain of origin, without this being a fully deterministic function of diagnosis or phenotype content.

\subsection{Neutral framing at full scale}

We re-ran the neutral prompt against all 1,115 originally-refused items to test whether removing the clinical role framing would recover more responses. Under this condition, 1,011 items (90.7\%) remained refused and 104 (9.3\%) were answered, closely matching the 8.0\% flip rate from the 25-item pilot. Table ~\ref{tab:rarebench_neutral} reports the full results under the neutral prompt. Scored on those 104 newly-answered items, Fable 5's Recall@1 (39.4\% {\scriptsize[30.6, 49.0]}) exceeds every other model's Recall@1 under the original prompt (20.9--27.7\%). This figure is computed on a self-selected subset, not a representative sample of the full benchmark, and should not be read as closing the gap with the other models at scale. It is, however, consistent with the pattern across all other benchmarks: wherever Fable 5 answers, it answers well.

\begin{table}[htbp]
\centering
\renewcommand{\arraystretch}{1.25}
\setlength{\tabcolsep}{6pt}
\resizebox{0.96\columnwidth}{!}{%
\begin{tabular}{p{3.7cm} p{3.6cm}}
\toprule
\textbf{Metric} & \textbf{Value} \\
\midrule
Model &
\textbf{Fable 5} \\
Originally refused pool &
1,115 \\
Still refused &
\textbf{1,011} {\scriptsize(90.7\%)} \\
Unparsed &
0 {\scriptsize(0.0\%)} \\
Scored responses &
\textbf{104} \\
\midrule
\rowcolor{fablerow}
R@1 & \textbf{39.42\%} {\scriptsize[30.6, 49.0]} \\
\rowcolor{fablerow}
R@5 & \textbf{45.19\%} {\scriptsize[36.0, 54.8]} \\
\rowcolor{fablerow}
R@10 & \textbf{47.12\%} {\scriptsize[37.8, 56.6]} \\
\bottomrule
\end{tabular}
}
\caption{RareBench performance and refusal under a neutral prompt. Fable~5 was tested only on the originally refused pool; brackets report 95\% Wilson confidence intervals.}
\label{tab:rarebench_neutral}
\vspace{-8pt}
\end{table}

\subsection{Refusal concentrates in basic-science and mechanism content}
\label{sec:basic_science}

\begin{figure}[h]
\centering
\includegraphics[width=0.48\textwidth]{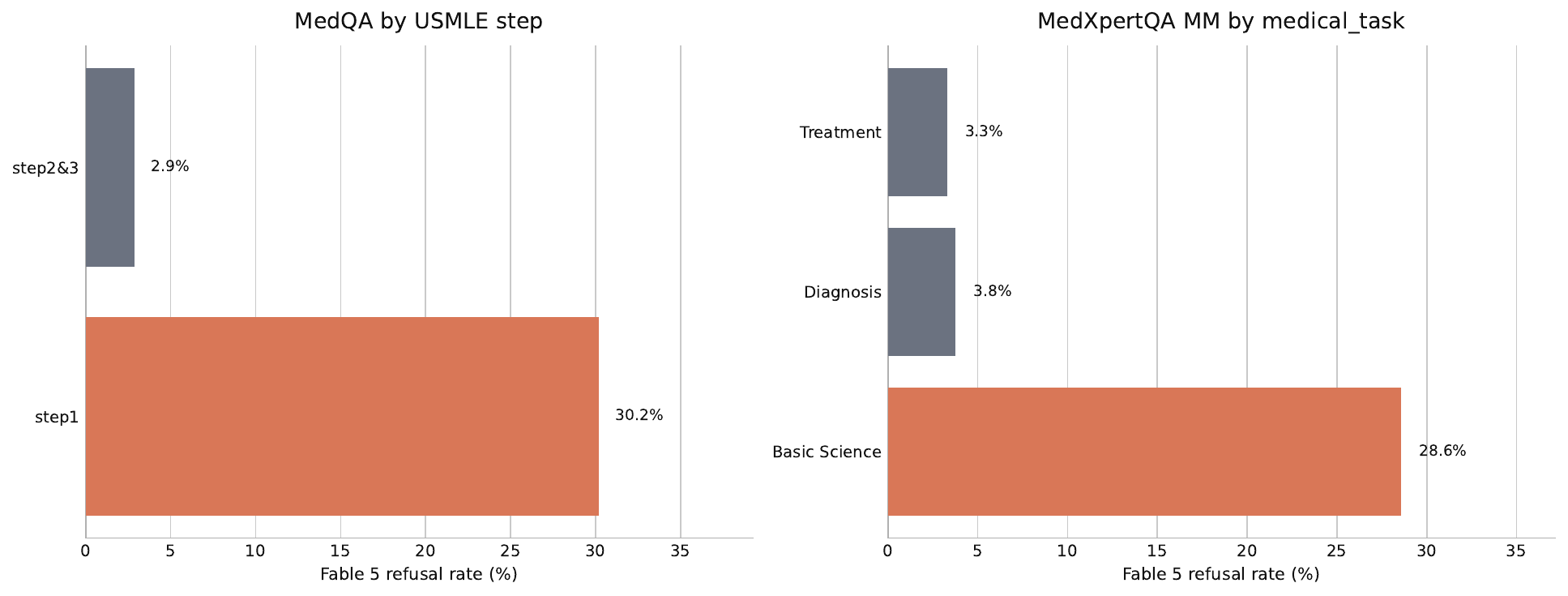}
\caption{Refusal rates by benchmark-specific content category: USMLE step (MedQA) and task type (MedXpertQA MM).}
\label{fig:refusal_rates}
\vspace{-2pt}
\end{figure}

Two independent benchmarks show the same underlying split. On MedQA, 92.3\% of Fable 5's refusals are Step 1 questions, preclinical, basic-science-oriented, against a 45.1\% baseline share of Step 1 among answered questions. On MedXpertQA MM, 62.5\% of refusals fall in the ``Basic Science'' category against a 13.6\% baseline share, using that benchmark's own independent labeling scheme. Figure ~\ref{fig:refusal_rates} shows refusal rates broken down by content category for both benchmarks. Close reading of refused MedXpertQA MM questions confirms the content pattern: many use a clinical-vignette wrapper around what is fundamentally a mechanism, anatomy, or physiology question, antibody structure, receptor pharmacology, cardiac electrophysiology, rather than a patient-specific diagnostic judgment.

\begin{table*}[t]
\centering
\renewcommand{\arraystretch}{1.35}
\setlength{\tabcolsep}{8pt}
\resizebox{0.92\textwidth}{!}{
\begin{tabular}{l c c c c}
\toprule
\textbf{Model}
  & \textbf{VQA-RAD} {\scriptsize(\textit{n}=179)}
  & \textbf{SLAKE} {\scriptsize(\textit{n}=250)}
  & \textbf{PathVQA scored} {\scriptsize(\textit{n}=100)}
  & \textbf{MedXpertQA scored} {\scriptsize(\textit{n}=368)} \\
\midrule
\rowcolor{fablerow}
\textbf{Fable 5}
  & \textbf{45.8\%} {\scriptsize[38.7, 53.1]}
  & \textbf{49.2\%} {\scriptsize[43.1, 55.4]}
  & \textbf{29.3\%} {\scriptsize[21.2, 38.9]}
  & \textbf{80.2\%} {\scriptsize[75.8, 83.9]} \\
Opus 4.8
  & 41.9\% {\scriptsize[34.7, 49.3]}
  & 47.0\% {\scriptsize[40.7, 53.4]}
  & 21.6\% {\scriptsize[16.1, 28.4]}
  & 62.3\% {\scriptsize[57.4, 66.9]} \\
Opus 4.6
  & 35.2\% {\scriptsize[28.6, 42.4]}
  & 38.0\% {\scriptsize[32.2, 44.2]}
  & 8.8\% {\scriptsize[5.4, 14.0]}
  & 55.0\% {\scriptsize[50.1, 59.8]} \\
GPT-5
  & 33.5\% {\scriptsize[27.0, 40.7]}
  & 40.4\% {\scriptsize[34.5, 46.6]}
  & 5.3\% {\scriptsize[2.8, 9.7]}
  & 67.2\% {\scriptsize[62.5, 71.7]} \\
\bottomrule
\end{tabular}
}
\caption{Multimodal benchmark performance. PathVQA and MedXpertQA MM report scored-subset accuracy after refusal exclusion. 95\% Wilson CIs in brackets. Opus 4.8 refused 7\% of SLAKE items}
\label{tab:multimodal_results}
\end{table*}

This pattern does not extend to RareBench. The mechanism-versus-presentation classification showed no discriminating power there, and the subset-level domain signal identified in Section~\ref{sec:rarebench_drivers} does not track the basic-science axis. We therefore report two related but distinct findings rather than a single unifying theory: across MedQA and MedXpertQA MM, refusal concentrates in basic-science and mechanism content; on RareBench, refusal concentrates by source dataset and disease domain. These two patterns are independently well-supported by the evidence. We were unable to unify them under a single mechanism, and we report that boundary explicitly rather than proposing an explanation the data does not support.

\subsection{Multimodal performance}

Table~\ref{tab:multimodal_results} reports accuracy across all four multimodal benchmarks. Fable 5 leads on three of the four. On VQA-RAD it scores 45.8\% {\scriptsize[38.7, 53.1]}, ahead of Opus 4.8 (41.9\%), Opus 4.6 (35.2\%), and GPT-5 (33.5\%). On MedXpertQA MM, scored on the 368 items Fable 5 answered, it reaches 80.2\% {\scriptsize[75.8, 83.9]}, a decisive margin over GPT-5 (67.2\% [62.5, 71.7]), with non-overlapping confidence intervals. On SLAKE, all four models cluster between 38.0\% and 49.2\% with substantially overlapping intervals; this benchmark does not discriminate among current systems. Opus 4.8 refused 7\% of SLAKE items, the only non-negligible refusal rate on this benchmark, without a clear effect on its relative ranking among the four models On PathVQA, Fable 5 refused 71 of 171 items (41.5\%); its scored-subset accuracy (29.3\% {\scriptsize[21.2, 38.9]}) places it above Opus 4.8 (21.6\%, no refusals) and the two lower-scoring models (Opus 4.6 7.6\%, GPT-5 5.3\%). Unlike the pattern in Section~\ref{sec:basic_science}, refused PathVQA items span both disease-process questions and purely descriptive ones, with no clear content pattern; we treat this as an open question.

\subsection{Generational trend and cross-model comparison}

Within the Claude family, the generational picture is consistently favorable to Fable 5 once refusal is separated from correctness. On every benchmark where Fable 5 answers, its accuracy matches or exceeds both predecessors; the multimodal gap on MedXpertQA MM (80.2\% vs.\ 62.3\% and 55.0\%) is the largest generational difference observed anywhere in this study. What changed release to release is not underlying capability but willingness: the refusal behavior documented in Sections~\ref{sec:refusal_raw}--\ref{sec:basic_science} did not exist in Opus 4.6 or Opus 4.8 at any comparable rate.

Against GPT-5, the comparison is benchmark -dependent but consistently favors Fable 5 on answered accuracy. On the legacy text benchmarks, GPT-5's raw scores exceed Fable 5's raw scores; Fable 5's scored-subset accuracy exceeds GPT-5's on both. On MedXpertQA in text and image form, Fable 5 leads outright. On VQA-RAD, Fable 5 leads; on SLAKE the two are statistically indistinguishable. On RareBench and PathVQA, GPT-5 attempts the task where Fable 5 largely does not, and that willingness gap, not a capability gap, is what GPT-5's higher raw scores on those benchmarks actually reflect.

The Appendix presents seven examples where Fable~5 answered correctly 
and all three other models did not, drawn from each benchmark, 
illustrating what Fable~5's reasoning looks like when it engages.

\section{Discussion}
\fontsize{11}{13}\selectfont
\subsection{What looked like capability decline was mostly refusal}

The raw scores alone tell a misleading story. On MedQA and PubMedQA, Fable 5 scores 10 to 15 percentage points below both predecessors and GPT-5 in raw accuracy, a result that, taken at face value, suggests the newest model in the family is weaker on basic medical knowledge than the ones before it. The refusal audit in Section~\ref{sec:refusal_raw} dismantles that reading. Once refused items are excluded from the denominator, Fable 5's accuracy on both benchmarks matches or exceeds every other model in this study. The apparent regression is an artifact of counting refusals as wrong answers, not evidence of reduced capability. What actually changed release to release is not what the model knows, it is what the model is willing to demonstrate.

This distinction matters for anyone using aggregate benchmark scores to track model progress. A single accuracy figure, collected without refusal tracking, would place Fable 5 below Opus 4.6 on benchmarks where it is in fact stronger. The right generational summary has two parts: Fable 5's demonstrated accuracy, on every benchmark where it answers, is competitive with or ahead of both predecessors; and a new pattern of selective refusal, absent in both Opus 4.6 and Opus 4.8 at any comparable rate, now sits between the model's capability and its measured performance.

\subsection{Refusal behavior and Anthropic's documented safeguards}

Anthropic's release documentation describes Fable 5's biology and chemistry classifier as targeting dual-use content capable of providing uplift to malicious actors - the company's own example is predicting how a genetic modification would affect viral capsid assembly in the context of pathogen design. The content that drove refusal in this study does not resemble that description. RareBench's most-refused cases are standard inborn metabolic disease presentations diagnosed from a routine phenotype list. MedQA's refused questions are preclinical board-exam material, anatomy, physiology, pharmacology, the same content medical students are examined on every year.

Two readings are possible. If the documented classifier is responsible for what we observed, it is engaging with content well outside its stated scope, at rates well above the sub-5\% session average Anthropic reports. If it is not responsible, the refusal behavior documented here does not correspond to any safeguard Anthropic has publicly described. We cannot determine which reading is correct from a black-box API evaluation, and we do not claim to. What this study can establish is that the observed refusal pattern, whatever its internal cause, does not behave as the documented mechanism predicts.

\subsection{The refusal pattern is not one phenomenon but two}

Refusal appears on five of the eight benchmarks in this study, at rates ranging from 8.0\% to 99.4\%. This is not an anomaly confined to one sensitive benchmark, but a recurring feature of Fable 5's behavior across legacy exam questions, specialist clinical images, and open-ended rare disease diagnosis alike.

Section~\ref{sec:basic_science} established that at least two distinguishable patterns are operating. The first runs across MedQA and MedXpertQA MM: refusal concentrates in basic-science and mechanism-level content, preclinical Step 1 questions, anatomy and physiology vignettes, receptor pharmacology, at rates far above their baseline share of each benchmark. Two independent benchmarks, using two entirely different official category schemes, converge on the same split. The second pattern is specific to RareBench: refusal tracks source dataset and disease domain, with inborn metabolic disease presentations refused near-universally and adult-onset autoimmune presentations answered at much higher rates. We tested directly whether the first pattern explains the second and found it does not, mechanism-dominant and presentation-dominant RareBench cases refused at statistically indistinguishable rates, and the classifier failed to track the subset-level pattern entirely. PathVQA presents a third signature that neither framework explains, with no content pattern emerging from manual inspection of its refused items.

We were unable to unify these under a single mechanism, and we report that boundary explicitly rather than proposing an explanation the data does not support. The Marfan syndrome counterexample in Section~\ref{sec:rarebench_drivers} subset, opposite refusal outcomes, illustrates that whatever governs this behavior operates at a finer grain than any variable we were able to measure.

\subsection{Practical implications}

Two findings bear directly on how a practitioner should approach deploying Fable 5 on biomedical tasks.

The first is a negative result, and we think it is the more useful one. Removing the clinical role framing from the RareBench prompt recovered only 9.3\% of originally-refused items at full scale. A practitioner who encounters this behavior and tries a simpler, less authoritative-sounding prompt will be disappointed roughly nine times out of ten. Prompt engineering is not a reliable mitigation for this refusal pattern, at least not via the most obvious intervention.

The second is consistently positive. Across every benchmark in this study, wherever Fable 5's response gets through, its accuracy is not lower than the other models' and is frequently higher. The primary constraint on its usefulness for biomedical tasks is willingness to engage, not competence once it does. That is a more tractable problem to design around, through case selection, fallback routing for refused items, or simply anticipating which question types are affected, than a genuine capability gap would be. A model that is highly accurate but selectively cautious is a different engineering challenge than a model that is simply wrong, and the two should not be treated the same way.

\subsection{Limits of what this paper establishes}

This paper measures behavior, not cause. We identify two real correlational patterns and one benchmark where no pattern emerged. We do not identify the underlying mechanism producing any of them. We have no access to Fable 5's training data, alignment procedure, or internal representations, and nothing in this paper's black-box methodology could establish a causal account in any case.

One earlier hypothesis, that refusal would concentrate on open-ended, free-text benchmarks rather than multiple-choice ones, the evidence directly contradicts. MedQA and MedXpertQA MM are both multiple-choice formats and both show substantial, categorically-patterned refusal. Format does not predict which benchmarks are affected. Content domain comes closer, but it does not generalize to RareBench or PathVQA. A full mechanistic account requires access this study does not have, most likely the model developer's own internal tooling. What this paper can establish, and does, is that refusal is a real, recurring, and benchmark-dependent behavior in Fable 5 that standard accuracy tables will not surface unless it is tracked and reported as a distinct outcome from the start.

\section{Related Work}
\fontsize{11}{13}\selectfont

Evaluating general-purpose language models on medical benchmarks began in earnest with Nori et al.\ \cite{Nori2023GPT4Medical}, who tested GPT-4 zero-shot on USMLE-style exams and treated calibration and safety-tuning costs as first-class findings alongside accuracy. Their methodology surfaced an early tension that the field has not resolved: safety-tuning cost GPT-4 three to five points of raw accuracy relative to its base model, an early sign that alignment and capability can pull against each other. Singhal et al.\ \cite{Singhal2023ClinicalKnowledge,Singhal2025ExpertMedicalQA} pushed the question further with Med-PaLM 2, adding human-rater judgment of factuality and potential harm on top of benchmark scores to ask whether a model a physician would actually trust. The most recent answer comes from Vishwanath et al.\ \cite{Vishwanath2026GeneralPurposeLLMs}, who found that GPT-5.2, Gemini 3.1 Pro, and Claude Opus 4.6 all outperformed OpenEvidence and UpToDate Expert AI on MedQA, HealthBench, and a real clinical query benchmark drawn from live physician use. Three years on, the answer to ``why test a general model at all'' remains the same one Nori et al.\ \cite{Nori2023GPT4Medical} implied: because it keeps winning.

The benchmarks in that original evaluation have not aged as well as the methodology. MedQA \cite{Jin2021MedQA} and PubMedQA \cite{Jin2019PubMedQA} are now near ceiling for frontier models, no longer separating one current system from another. MedXpertQA \cite{Zuo2025MedXpertQA} and RareBench \cite{Chen2024RareBench} were built to restore what those benchmarks lost. Zuo et al.\ \cite{Zuo2025MedXpertQA} move from four-option licensing questions to ten-option specialist-board questions across 17 specialties, with published top-model performance as low as 46\% on the harder reasoning subset. Chen et al.\ \cite{Chen2024RareBench} go further, replacing multiple-choice recognition with open-ended diagnosis generation, where standalone models land between 14.6\% and 40\% Recall@1. Both are treated as primary evidence benchmarks here for exactly that reason.

For image-based evaluation, we draw on VQA-RAD \cite{Lau2018VQARAD}, SLAKE \cite{Liu2021SLAKE}, PathVQA \cite{He2020PathVQA}, and the multimodal subset of MedXpertQA \cite{Zuo2025MedXpertQA}. These benchmarks cover radiology, multi-modality clinical imaging, and histopathology, domains where real biomedical questions increasingly arrive with an image attached. We exclude closed-ended yes/no subsets from all three VQA datasets, as binary questions no longer require a model to produce an answer, only confirm one. MedXpertQA's multimodal subset pairs expert-level exam questions with real clinical images and patient records, making it the hardest task in this phase and the one most directly analogous to specialist clinical reasoning.

What none of this prior work does is examine more than one point in a single model's lineage. Every study above is a snapshot, one generation compared against another at a single moment in time. The generational trace in this paper, running Opus 4.6, Opus 4.8, and Fable 5 under identical conditions, addresses that gap directly.

Reliable scoring is a prerequisite for any of these comparisons to hold. Grading open-ended responses with another language model introduces the judge's own biases and failure modes into the result, a risk that is especially damaging when the central claim of a paper is that one model outperforms another. A February 2026 audit of MedCalc-Bench illustrates the concrete cost of this kind of validity problem, even outside the LLM-as-judge setting specifically: Krohn-Grimberghe \cite{KrohnGrimberghe2026MedCalcBenchAudit} identified and corrected more than twenty formula and runtime errors in the benchmark's own calculator implementations, and separately showed that GPT-5.2-Thinking reaches 95--97\% accuracy once given the calculator specification at inference time, with the remaining errors attributable to ground-truth issues rather than genuine reasoning failures, evidence that the benchmark's apparent difficulty reflected implementation bugs and formula memorization more than clinical reasoning. Every benchmark in this study is scored against a deterministic ground truth, with no model judgment anywhere in the pipeline.

\section{Conclusion}
\fontsize{11}{13}\selectfont

Claude Fable 5's biomedical capability, measured on what it actually answers, is strong, stronger than its raw scores in this paper suggest. On text benchmarks, its scored-subset accuracy meets or exceeds every other model evaluated, including on MedQA and PubMedQA, where its raw numbers alone would have suggested a regression from its own predecessors. On multimodal benchmarks, it leads outright on three of four, and on MedXpertQA MM, arguably the hardest, most specialist-oriented benchmark in this study, it beats every other model by a wide, non-overlapping margin, even with refusal included in the denominator. The gap between that underlying capability and Fable 5's raw performance is \textbf{refusal, not competence}. Refusal ranges from 8.0\% to 99.4\% depending on the benchmark, does not appear in either Claude predecessor evaluated here, and follows at least two distinguishable patterns rather than one uniform cause. That gap is this paper's central finding, but it should not overshadow the capability finding underneath it: when Fable 5 engages with a biomedical question, it is, on the evidence in this study, among the strongest models tested.

\section{Limitations}
\fontsize{11}{13}\selectfont

\textbf{Domain and prompting scope.} This study covers eight benchmarks across general clinical knowledge, specialist reasoning, rare disease diagnosis, and four imaging domains, evaluated single-shot with direct-answer prompting, following the Nori at al.\cite{Nori2023GPT4Medical} template. It does not extend to other biomedical domains, genomics, pharmacology, mental health, among others, where refusal behavior may differ, and it does not test chain-of-thought prompting, which was excluded primarily on cost grounds and might affect both accuracy and refusal rate. Both are left to future work.

\textbf{Scope of comparison.} We compare against one external model at one point in time, with no human physician baseline; this was a deliberate scope decision, not an oversight.

\textbf{Single-run evaluation.} Every figure reflects one API call per item. Refusal-rate estimates are point estimates from a single run, not stability-tested across repeated queries.

\vspace{0.5cm}

\balance
{
\footnotesize
\setlength{\bibitemsep}{2pt}
\printbibliography
}

\newpage
\onecolumn
\appendix

\section{Appendix}
\fontsize{11}{13}\selectfont

\section*{A. Appendix}

Table~\ref{tab:unique_correct} reports, for each benchmark, how many items Fable 5 answered correctly while Opus 4.6, Opus 4.8, and GPT-5 all missed. This is not a claim about aggregate accuracy, Tables~\ref{tab:text_results}, \ref{tab:rare_text_results}, and~\ref{tab:multimodal_results} already cover that, but a qualitative look at what these items contain. Figures~A1 through A7 present one example from each benchmark where Fable 5 answered correctly and all three other models did not, showing the question, Fable 5's response, and the ground-truth answer. The examples give a concrete sense of what Fable 5's reasoning looks like when it engages with a question, complementing the numerical case made throughout Section~\ref{sec:results}. The seven examples are chosen for clarity, not difficulty or representativeness, and should not be read as a random sample. RareBench contributes none, since only 6 items were scored at all under the primary prompt (Section~\ref{sec:rarebench}).

\begin{table}[h]
\centering
\small
\renewcommand{\arraystretch}{1.25}
\begin{tabular}{l c c}
\toprule
\textbf{Benchmark} & \textbf{Uniquely correct} & \textbf{Total scored} \\
\midrule
MedQA & 7 & 1,273 \\
PubMedQA & 18 & 1,000 \\
MedXpertQA Text & 134 & 2,450 \\
RareBench & 0 & 6 \\
VQA-RAD & 12 & 179 \\
PathVQA & 14 & 171 \\
SLAKE & 9 & 250 \\
MedXpertQA MM & 27 & 368 \\
\bottomrule
\end{tabular}
\caption{Items where Fable 5 answered correctly and Opus 4.8, Opus 4.6, and GPT-5 all answered incorrectly. RareBench's zero count is consistent with (Section~\ref{sec:rarebench}): only 6 items were scored at all under the primary prompt.}
\label{tab:unique_correct}
\end{table}

\tcbset{appendixstyle/.style={
  enhanced,
  colback=gray!2,
  colframe=gray!30,
  colbacktitle=gray!15,
  coltitle=black,
  boxrule=0.5pt,
  arc=3pt,
  outer arc=3pt,
  lefttitle=8pt,
  toptitle=5pt,
  bottomtitle=5pt,
  left=8pt,
  right=8pt,
  top=7pt,
  bottom=7pt,
  fonttitle=\bfseries\small,
}}

\begin{tcolorbox}[appendixstyle, title=MedQA \#677 --- Fable 5 uniquely correct]
\small
\textbf{Question:} A 2-month-old infant is brought to the emergency department by her mother due to a fall. Tearfully, the patient's mother describes witnessing her child fall from the changing table. The patient was born prematurely at 36 weeks estimated gestational age via vacuum-assisted vaginal delivery. The patient is afebrile. Her vital signs include: blood pressure of 94/60 mm Hg, pulse 200/min, and respiratory rate 70/min. Physical examination reveals a subconjunctival hemorrhage in the left eye and multiple bruises on the chest and back. Which of the following is the best initial step in management of this patient's condition?

\medskip
\textbf{Answer Choices:}\\
(A) CT scan of the head \quad (B) Involvement of social services\\
(C) Family counseling \quad (D) Rule out medical conditions

\medskip
\textbf{Correct answer:} A

\medskip
\textbf{Fable 5's response:}\\
\textit{This 2-month-old presents with findings highly concerning for non-accidental trauma (child abuse): multiple bruises on the chest and back in a non-mobile infant, subconjunctival hemorrhage, and an injury pattern inconsistent with the reported mechanism. More importantly, the infant is hemodynamically unstable, with marked tachycardia (200/min) and tachypnea (70/min), raising concern for intracranial hemorrhage. The priority in suspected child abuse is always medical stabilization first --- social services involvement (B) is legally mandated but comes after the child is medically stabilized.}

\medskip
\textbf{ANSWER: A}
\end{tcolorbox}

\vspace{4pt}
{\footnotesize\textbf{Figure A1.} Fable 5 correctly sequences medical stabilization before mandated reporting in a case of suspected non-accidental trauma, while Opus 4.6, Opus 4.8, and GPT-5 each selected a different option.}

\vspace{1.2cm}

\begin{tcolorbox}[appendixstyle, title=PubMedQA \#23497210]
\small
\textbf{Question:} Are women with major depression in pregnancy identifiable in population health data?

\medskip
\medskip
\textbf{Context (abridged):} A population-based linkage study compared two methods of identifying maternal depression during pregnancy --- antidepressant dispensing records and hospital admission comorbidity records --- across 96{,}698 births in Western Australia. Only 899 of 6{,}596 identified pregnancies (13.6\%) were captured by \emph{both} data sources; 80\% of women dispensed an antidepressant had no depression comorbidity recorded on their hospital records.

\medskip
\textbf{Correct answer:} no

\medskip
\medskip
\textbf{Fable 5's response:}\\
\textit{The study found very poor concordance between the two data sources: only 13.6\% of pregnancies were captured by both. Notably, 80\% of women dispensed antidepressants had no depression comorbidity recorded on their hospital records. Furthermore, the capture-recapture estimate suggests true prevalence (\textasciitilde16\%) is far higher than what either dataset identified individually. This indicates substantial under-ascertainment.}

\medskip
\medskip
\textbf{ANSWER: no}
\end{tcolorbox}

\vspace{4pt}
{\footnotesize\textbf{Figure A2.} Fable 5 correctly reasons from cross-source discordance and a capture-recapture estimate to conclude that neither administrative dataset reliably identifies maternal depression alone.}

\vspace{1.2cm}

\begin{tcolorbox}[appendixstyle, title=MedXpertQA Text \#Text-20]
\small
\textbf{Question:} A 57-year-old male with a history of IV drug use, hepatitis C, atrial fibrillation, cirrhosis, alcohol dependence, obesity, and depression presents with a one-week history of weakness and fever. He is admitted and started on piperacillin-tazobactam for suspected spontaneous bacterial peritonitis. By day 4, he is afebrile and improving, but nurses report bleeding from IV sites and new bruising. What is the most likely explanation for this patient's condition?

\medskip
\medskip
\textbf{Answer Choices:}\\
(A) Fibrinolysis activation due to liver failure \quad
(B) Vitamin K deficiency due to liver dysfunction\\
(C) Diffuse activation of the coagulation cascade \quad
(D) Disseminated intravascular coagulation (DIC)\\
(E) Platelet dysfunction secondary to sepsis \quad
(F) Decreased production of coagulation factors\\
(G) Immune-mediated platelet destruction \quad
(H) Hypersplenism causing thrombocytopenia\\
(I) Decreased metabolism of an anticoagulant \quad
(J) Bacterial destruction of clotting factors

\medskip
\medskip
\medskip
\textbf{Correct answer:} J

\medskip
\textbf{Fable 5's response:}\\
\textit{This patient has atrial fibrillation, for which he is almost certainly anticoagulated with warfarin. The key clue is timing: bleeding developed on hospital day 4, after several days of broad-spectrum antibiotics. These antibiotics eradicate colonic flora that synthesize vitamin K, reducing synthesis of factors II, VII, IX, and X --- an effect that dramatically potentiates warfarin's anticoagulant action. DIC (C, D) would be expected with worsening sepsis, but this patient is improving; piperacillin-tazobactam does not significantly inhibit warfarin metabolism (I), unlike agents such as metronidazole or fluconazole.}

\medskip
\medskip
\textbf{ANSWER: J}
\end{tcolorbox}

\vspace{4pt}
{\footnotesize\textbf{Figure A3.} Fable 5 correctly identifies gut-flora-mediated vitamin K depletion potentiating an unstated pre-existing warfarin regimen, using the day-4 timing to rule out DIC and platelet-related causes despite a septic-sounding presentation. Opus 4.6, Opus 4.8, and GPT-5 each selected a different option among ten choices.}

\begin{tcolorbox}[appendixstyle, title=VQA-RAD \#vqa\_rad\_26 --- Fable 5 uniquely correct]
\small
\includegraphics[width=0.32\linewidth]{vqa_rad_26.jpg}

\medskip
\textbf{Question:} What type of MRI modality is used?

\medskip
\textbf{Ground-truth answer:} DWI (diffusion weighted)

\medskip
\textbf{Fable 5's response:}\\
\textit{The image shows a diffusion-weighted MRI (DWI) sequence, evident from the bright hyperintense signal in the left MCA territory consistent with acute infarction.}

\medskip
\textbf{ANSWER: DWI (diffusion weighted imaging)}
\end{tcolorbox}

\vspace{4pt}
{\footnotesize\textbf{Figure A4.} Fable 5 correctly identifies the specific MRI sequence, not just the imaging modality class, reasoning from the acute-infarction signal pattern. Opus 4.6, Opus 4.8, and GPT-5 each answered incorrectly.}

\vspace{1.2cm}

\begin{tcolorbox}[appendixstyle, title=PathVQA \#path\_vqa\_58 --- Fable 5 uniquely correct]
\small
\includegraphics[width=0.38\linewidth]{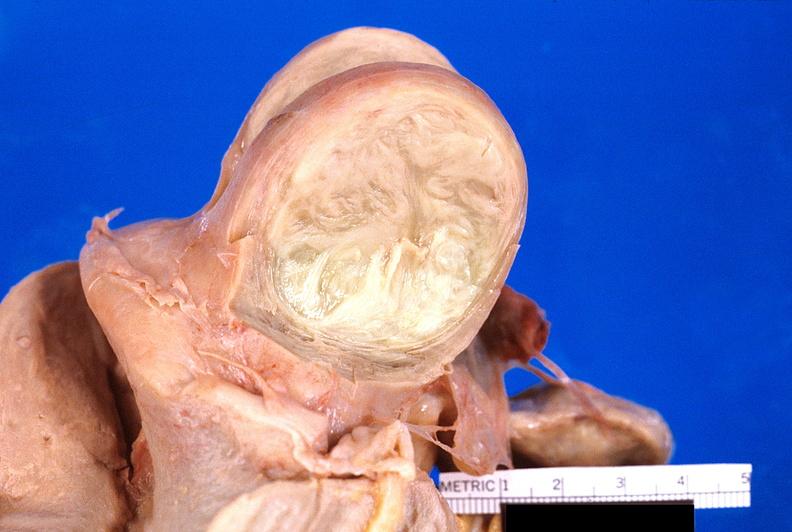}

\medskip
\textbf{Question:} What does this image show?

\medskip
\textbf{Ground-truth answer:} uterus

\medskip
\textbf{Fable 5's response:}\\
\textit{This image shows a gross specimen of a uterus with a large leiomyoma (fibroid), demonstrating the characteristic whorled, firm, white-tan cut surface of the tumor.}

\medskip
\textbf{ANSWER: uterus with leiomyoma}
\end{tcolorbox}

\vspace{4pt}
{\footnotesize\textbf{Figure A5.} Fable 5 identifies both the organ and the specific pathological finding, describing the gross morphological features supporting the diagnosis. Opus 4.6, Opus 4.8, and GPT-5 each answered incorrectly.}

\vspace{1.2cm}

\begin{tcolorbox}[appendixstyle, title=SLAKE \#slake\_245 --- Fable 5 uniquely correct]
\small
\includegraphics[width=0.32\linewidth]{slake_245.jpg}

\medskip
\textbf{Question:} What color does the right lung show in the picture?

\medskip
\textbf{Ground-truth answer:} black

\medskip
\textbf{Fable 5's response:}\\
\textit{The right lung appears dark (black) on this chest X-ray, as air-filled lung tissue is radiolucent.}

\medskip
\textbf{ANSWER: black}
\end{tcolorbox}

\vspace{4pt}
{\footnotesize\textbf{Figure A6.} Fable 5 grounds its answer in an imaging physics principle (radiolucency of air-filled tissue) rather than surface-level pattern matching. Opus 4.6, Opus 4.8, and GPT-5 each answered incorrectly.}

\vspace{1.2cm}

\begin{tcolorbox}[appendixstyle, title=MedXpertQA MM \#MM-1938 --- Fable 5 uniquely correct]
\small
\includegraphics[width=0.2\linewidth]{MM-1938-a.png}\hspace{0.5cm}\includegraphics[width=0.22\linewidth]{MM-1938-b.png}

\medskip
\textbf{Question:} In a 58-year-old female patient, what is the MOST likely diagnosis?

\medskip
\textbf{Answer Choices:}\\
(A) Osteoarthritis \quad (B) Hemophiliac arthropathy \quad (C) Gout\\
(D) Rheumatoid arthritis \quad (E) Amyloid deposition

\medskip
\textbf{Correct answer:} D

\medskip
\textbf{Fable 5's response:}\\

\medskip
\textbf{ANSWER: D}
\end{tcolorbox}

\vspace{4pt}
{\footnotesize\textbf{Figure A7.} Fable 5 correctly selects rheumatoid arthritis from a five-option specialist-level differential grounded in the accompanying clinical images. Opus 4.6, Opus 4.8, and GPT-5 each selected a different option. No reasoning text accompanies this answer, consistent with the direct-answer, no-CoT prompting protocol used throughout this study (Section~\ref{sec:prompting})

\begin{tcolorbox}[appendixstyle, title=SLAKE \#slake\_229 --- Fable 5 uniquely correct]
\small
\includegraphics[width=0.32\linewidth]{slake_229.jpg}

\medskip
\textbf{Question:} In what plane is this image scanned?

\medskip
\textbf{Ground-truth answer:} Transverse Plane

\medskip
\textbf{Fable 5's response:}\\
\textit{This is a CT image showing a cross-sectional (transverse) view of the abdomen at the level of the liver and spine.}

\medskip
\textbf{ANSWER: Axial (transverse) plane}
\end{tcolorbox}

\vspace{4pt}
{\footnotesize\textbf{Figure A8.} Fable 5 correctly identifies the imaging orientation as the axial (transverse) plane and supports its answer by recognizing the scan as a cross-sectional abdominal CT image. Opus 4.6, Opus 4.8, and GPT-5 each answered incorrectly.}

\end{document}